\documentclass[letterpaper, 10 pt, conference]{ieeeconf}  

\IEEEoverridecommandlockouts                              

\usepackage{amsmath} 
\usepackage{amssymb}  
\usepackage[usenames,dvipsnames]{color}
\usepackage{graphicx}
\usepackage{ifthen}
\usepackage[normalem]{ulem}
\usepackage{url}

\title{\LARGE \bf
Deep Multimodal Embedding: 
Manipulating Novel Objects \\ with Point-clouds, Language and Trajectories
}

\author{Jaeyong Sung$^{1,2}$, Ian Lenz$^{1}$ and Ashutosh Saxena$^{3}$
\thanks{$^{1}$Department of Computer Science, Cornell University.
        $^{2}$Department of Computer Science, Stanford University.
        $^{3}$Brain Of Things, Inc.
        {\tt\small \{jysung,ianlenz,asaxena\}@cs.cornell.edu}}%
}

\newcommand{\citet}[1]{\cite{#1}}

\newcommand{\todo}[1]{\textcolor{blue}{[\textbf{#1}]}}

\newcommand{\argmax}{\arg\!\max}

\newcommand{\header}[1]{\smallskip\noindent\textbf{#1}}

\newboolean{include-notes}
\setboolean{include-notes}{true}
\newcommand{\ilnote}[1]{\ifthenelse{\boolean{include-notes}}%
 {\textcolor{Bittersweet}{[\textbf{IL: #1}]}}{}}
\newcommand{\jaenote}[1]{\ifthenelse{\boolean{include-notes}}%
 {\textcolor{PineGreen}{[\textbf{JS: #1}]}}{}}

\definecolor{darkgreen}{rgb}{0.0, 0.2, 0.13}

\begin{document}
\maketitle

\begin{abstract}
A robot operating in a real-world environment needs to perform reasoning 
over a variety of sensor modalities such as vision, language and motion trajectories. 
However, it is extremely challenging to 
manually design features relating such disparate modalities.
In this work, we introduce an algorithm that learns to embed 
point-cloud, natural language, and manipulation trajectory data 
into a shared embedding space with a deep neural network.
To learn semantically meaningful spaces throughout our network,
we use a loss-based margin to bring embeddings of relevant pairs closer together
while driving less-relevant cases from different modalities further apart. 
We use this both to pre-train its lower layers and fine-tune our final embedding space,
leading to a more robust representation.
We test our algorithm on the task of manipulating novel objects and appliances
based on prior experience with other objects.
On a large dataset, we achieve significant improvements in both accuracy and inference time 
over the previous state of the art. 
We also perform end-to-end experiments on a PR2 robot
utilizing our learned embedding space.
\end{abstract}


  \abovedisplayskip 3.0pt plus2pt minus2pt%
 \belowdisplayskip \abovedisplayskip
\renewcommand{\baselinestretch}{0.97}

\newenvironment{packed_enum}{
\begin{enumerate}
  \setlength{\itemsep}{0pt}
  \setlength{\parskip}{0pt}
  \setlength{\parsep}{0pt}
}
{\end{enumerate}}

\newenvironment{packed_item}{
\begin{itemize}
  \setlength{\itemsep}{0pt}
  \setlength{\parskip}{0pt}
  \setlength{\parsep}{0pt}
}{\end{itemize}}

\newlength\savedwidth
\newcommand\whline[1]{\noalign{\global\savedwidth\arrayrulewidth
                               \global\arrayrulewidth #1} %
                      \hline
                      \noalign{\global\arrayrulewidth\savedwidth}}

\newlength{\sectionReduceTop}
\newlength{\sectionReduceBot}
\newlength{\subsectionReduceTop}
\newlength{\subsectionReduceBot}
\newlength{\abstractReduceTop}
\newlength{\abstractReduceBot}
\newlength{\captionReduceTop}
\newlength{\captionReduceBot}
\newlength{\subsubsectionReduceTop}
\newlength{\subsubsectionReduceBot}
\newlength{\headerReduceTop}
\newlength{\figureReduceBot}

\newlength{\horSkip}
\newlength{\verSkip}

\newlength{\equationReduceTop}

\newlength{\figureHeight}
\setlength{\figureHeight}{1.7in}

\setlength{\horSkip}{-.09in}
\setlength{\verSkip}{-.1in}

\setlength{\figureReduceBot}{0in}
\setlength{\headerReduceTop}{0in}
\setlength{\subsectionReduceTop}{-0.03in}
\setlength{\subsectionReduceBot}{-0.05in}
\setlength{\sectionReduceTop}{0in}
\setlength{\sectionReduceBot}{-0.01in}

\setlength{\subsubsectionReduceTop}{0in}
\setlength{\subsubsectionReduceBot}{0in}
\setlength{\abstractReduceTop}{0in}
\setlength{\abstractReduceBot}{0in}

\setlength{\equationReduceTop}{0in}

\setlength{\captionReduceTop}{0in}
\setlength{\captionReduceBot}{0in}

\section{Introduction}

Consider a robot manipulating a new appliance in a home kitchen, 
\textit{e.g.} the toaster in Figure~\ref{fig:system_overview}. 
The robot must use the combination of its observations
of the world and natural language instructions
to infer how to manipulate objects. 
Such ability to fuse information from different input modalities
and map them to actions is extremely useful 
to many applications of household robots \cite{sung_robobarista_2015},
including assembling furniture, cooking recipes, and many more.

Even though similar concepts might appear very differently in
different sensor modalities, 
humans are able to understand that they map to the same concept.
For example, when asked to ``turn the knob counter-clockwise'' on a toaster,
we are able to correlate 
the instruction language and the appearance of a knob on a toaster
with the motion to do so.
We also associate this concept more
closely with a motion which would
incorrectly rotate in the opposite
direction than with, for example,
the motion to press the toaster's handle
downwards.
There is strong evidence that humans are able to correlate between different modalities
through \emph{common representations} \cite{erdogan2014transfer}.

Obtaining a good common representation between different modalities is
challenging for two main reasons. First, each modality might intrinsically have
very different statistical properties --- for example, here
our trajectory
representation is inherently dense, while our 
representation of
language is naturally sparse. This makes it challenging to apply algorithms
designed for unimodal data.
Second, even with expert knowledge, it is extremely challenging to design
joint features between such disparate modalities. 
Designing features which map different sensor inputs and actions to the same
space, as required here, is particularly challenging.

\begin{figure}
  \begin{center}
    \includegraphics[width=0.99\columnwidth]{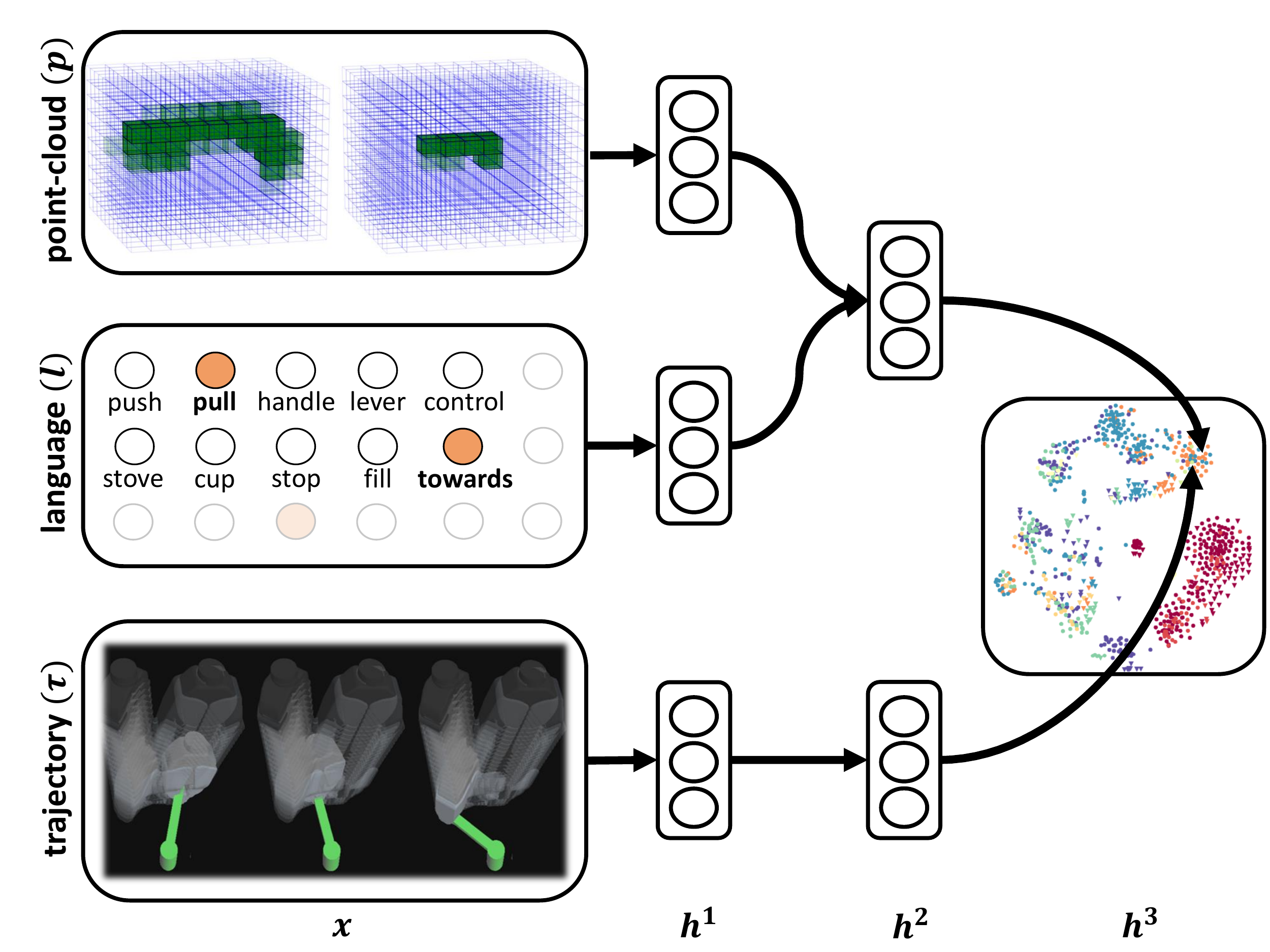}
  \end{center}
  \vskip -0.15in
  \caption{
    \textbf{Deep Multimodal Embedding:} 
    Our deep neural network learns 
    to embed both point-cloud/natural language instruction combinations and
    manipulation trajectories in the same semantically meaningful space,
    where distance represents the relevance of embedded data.
    }
  \label{fig:main_fig}
\end{figure}

In this work, 
we use a deep neural network to learn a shared embedding 
between the pairing of object parts in the environment with natural language instructions, 
and manipulation trajectories (Figure~\ref{fig:main_fig}). 
This means that all three modalities are projected to the
\emph{same} feature space.
We introduce an algorithm that learns to pull semantically
similar environment/language pairs and their corresponding trajectories 
to the same
regions, and push environment/language pairs away from irrelevant trajectories
based on how irrelevant they are.
Our algorithm allows for efficient inference because,
given a new instruction and point-cloud,
we need only find the nearest trajectory to the projection of this pair
in the learned embedding space, which can be done using fast 
nearest-neighbor algorithms \cite{flann_pami_2014}.

In the past, deep learning methods have shown impressive results
for learning features for a wide variety of domains 
\cite{krizhevsky2012imagenet,socher2011semi,hadsell2008deep}
and even learning cross-domain embeddings
\cite{srivastava2012multimodal}. 
In contrast to these existing methods, here we present a new
pre-training algorithm for initializing networks to be used for
joint embedding of different modalities. 
Our algorithm trains each layer to map similar
cases to similar areas of its feature space, as opposed to other
methods which either perform variational learning \cite{hinton2006reducing}
or train for reconstruction \cite{SAE}.

In order to validate our approach, we test our model on a large manipulation 
dataset, the Robobarista Dataset \cite{sung_robobarista_2015},
which focuses on learning to infer manipulation trajectories for \textit{novel} objects. 
We also present results on a series of experiments on a PR2 robot,
showing that our algorithm is able to 
successfully 
manipulate several different objects.
In summary, the key contributions of this work are:
\begin{packed_item}
\item We present an algorithm which learns a semantically meaningful embedding space 
by enforcing a varying and loss-based margin. 
\item We present an algorithm for unsupervised pre-training of multi-modal features 
to be used for  embedding which outperforms standard pre-training algorithms \cite{SAE}.
\item We present a novel approach to manipulation 
via 
embedding multimodal data with deep neural networks.
\item Our approach allows for fast inference
of manipulation trajectories for novel objects,
roughly $170$x faster than previous work \cite{sung_robobarista_2015} while also improving accuracy
on a large manipulation dataset \cite{sung_robobarista_2015}.
\end{packed_item}

\section{Related Work}
\vspace*{\sectionReduceBot}

\vspace*{\subsectionReduceTop}
\subsection{Metric Embedding}
\vspace*{\subsectionReduceBot}

Several works in machine learning make use of the power of shared embedding spaces.
Large margin nearest neighbors (LMNN) \cite{weinberger2005distance} learns a max-margin Mahalanobis distance 
for a unimodal input feature space.
\citet{weston2011wsabie} learn linear mappings from image and language features to a common embedding 
space for automatic image annotation. \citet{moore2012playlist} learn to map songs and natural language
tags to a shared embedding space.
However, these approaches learn only a shallow, linear mapping from input features, whereas here we learn
a deep non-linear mapping which is less sensitive to input representations.

\vspace*{\subsectionReduceTop}
\subsection{Deep Learning}
\vspace*{\subsectionReduceBot}

\header{Multimodal data:}
A deep neural network has been used to learn features of video and audio \cite{ngiam2011multimodal}.
With a generative learning, a network can be robust to missing modalities 
at inference time \cite{sohn2014improved}.
In these works, similar to \citet{sung_robobarista_2015}, 
a single network takes all modalities as inputs,
whereas here we perform joint embedding of multiple modalities using multiple networks.

\header{Joint embedding:}
Several works use deep networks for joint embedding between different feature spaces.
For translation, a joint feature space is learned from different languages \cite{mikolov2013translation};
for annotation and retrieval, images and natural language ``tags'' 
are mapped to the same space \citet{srivastava2012multimodal}.
We present a new pre-training algorithm for
embedding spaces and show that it outperforms the
conventional methods used in these works.



A deep network is also used as metric learning 
for the face verification task \cite{hu2014discriminative},
which enforces a constant margin between
distances among inter-class objects and among intra-class objects,
similar to LMNN \cite{weinberger2005distance}.
In Sec.~\ref{sec:results}, we show that our approach, which uses a
loss-dependent variable margin, produces better results for our problem.

\vspace*{\subsectionReduceTop}
\subsection{Robotic Manipulation}
\vspace*{\subsectionReduceBot}

Many works in robotic manipulation focus on task-specific
manipulation with \emph{known} objects --- for example, folding
towels \cite{miller2012geometric}, baking
cookies \cite{bollini2011bakebot},
or planar contact manipulation \cite{Koval_2015_7998}. 
Others \cite{sung_synthesizingsequences_2014,misra2014tellme} 
focus on sequencing manipulation tasks or choosing when to switch skills \cite{Kappler-RSS-15}, assuming 
manipulation primitives such as \textit{pour} are available.
For novel objects, affordances are predicted and associated motions are applied \cite{kroemer2012kernel}. 
Instead, similar to \cite{sung_robobarista_2015}, we skip intermediate
representations and 
directly generalize to novel objects.

A few recent works use deep learning approaches for 
robotic manipulation. 
Deep networks have been used to detect stable grasps
from RGB-D data \cite{kappler2015leveraging,lenz2013deep}.
\citet{levine2015manipulation} use a Gaussian mixture
model to learn system dynamics, then use these to learn 
a manipulation policy using a deep network. 
\citet{lenz2015deepmpc} use a deep network to learn system dynamics for real-time model-predictive control. 
Both these works
focus on learning low-level input-output controllers.
Here, we instead focus on inferring full 6-DoF trajectories, which such controllers could then be used
to follow.

Sung et al. \citet{sung_robobarista_2015} perform
object part-based  transfer of manipulation trajectories for novel objects 
and introduces a large manipulation dataset  
including objects like epresso machine and urinal.
We primarily test our algorithm on this dataset.
In Sec.~\ref{sec:results}, we show that our approach gives better
accuracy than this prior work, while also running $171$x faster.



\section{Overview}
\vspace*{\sectionReduceBot}

The main challenge of our work is to learn a model which maps three disparate
modalities --- point-clouds, natural language, and trajectories --- to a single
semantically meaningful space.
In particular, we focus on point-clouds of object parts,
natural language instructing manipulation of different objects,
and trajectories that would manipulate these objects.

We introduce a method that learns a common point-cloud/language/trajectory embedding space 
in which the projection of a task (point-cloud/language combination)
should have a higher similarity to projections of relevant trajectories
than task-irrelevant trajectories.
Among these irrelevant trajectories, some might be less relevant than others,
and thus should be pushed further away.

For example, given a door knob, that needs to be grasped normal to the door surface, with
an instruction to rotate it clockwise, 
a trajectory that correctly approaches the door knob but rotates counter-clockwise
should have higher similarity to the task than one
which approaches the knob from a completely incorrect angle and does not execute any rotation.

We learn non-linear embeddings using a deep learning approach,
as shown in Fig.~\ref{fig:main_fig},
which maps raw data from these three different modalities to a joint embedding space.
Prior to learning a full joint embedding of all three modalities,
we pre-train embeddings of subsets of the modalities 
to learn semantically meaningful embeddings for these modalities.

We show in Sec.~\ref{sec:systems}
that a learned joint embedding space can be efficiently used 
for finding an appropriate manipulation trajectory
for objects with  natural language instructions.

\subsection{Problem Formulation}
\vspace*{\subsectionReduceBot}

Given tuples of a scene $p \in \mathcal{P}$, 
a natural language instruction $l \in \mathcal{L}$
and an end-effector trajectories $t \in \mathcal{T}$,
our goal is to learn a joint embedding space 
and two different mapping functions
that map to this space---one from a point-cloud/language pair 
and the other from a trajectory.

More formally, we want to learn $\Phi_{\mathcal{P},\mathcal{L}}(p,l)$ and
$\Phi_{\mathcal{T}}(\tau)$ which map to a joint feature space $\mathbb{R}^M$:
\begin{align*}
\Phi_{\mathcal{P},\mathcal{L}}(p,l)&: (\mathcal{P},\mathcal{L}) \rightarrow \mathbb{R}^M  \\
\Phi_{\mathcal{T}}(\tau)&: \mathcal{T} \rightarrow \mathbb{R}^{M}
\end{align*}
The first, $\Phi_{\mathcal{P},\mathcal{L}}$, which maps point-clouds and languages, 
is defined as a combination of two mappings. The first of these maps to a joint 
point-cloud/language space $\mathbb{R}^{N_{2,pl}}$ ---
$\Phi_{\mathcal{P}}(p):\mathcal{P} \rightarrow \mathbb{R}^{N_{2,pl}}$ and
$\Phi_{\mathcal{L}}(l):\mathcal{L} \rightarrow \mathbb{R}^{N_{2,pl}}$.
$N_{2,pl}$ represents the size of dimensions $p,l$ are embedded jointly.
Once each is mapped to $\mathbb{R}^{N_{2,pl}}$, 
this space is then mapped to the joint space shared with trajectory information: 
$\Phi_{\mathcal{P},\mathcal{L}}(p,l): ((\mathcal{P}, \mathcal{L}) \rightarrow \mathbb{R}^{N_{2,pl}}) \rightarrow \mathbb{R}^M$.

\section{Learning Joint Point-cloud/Language/Trajectory Model}
\label{sec:DME}

In our joint feature space, proximity between two mapped points 
should reflect how relevant two data-points are to each other,
even if they are from completely different modalities.
We train our network to bring demonstrations that manipulate a given object 
according to some language instruction closer to the mapped point for that object/instruction pair,
and to push away demonstrations that would not correctly manipulate that object.
Trajectories which have no semantic relevance to the object are pushed much further 
away than trajectories that have some relevance, 
even if the latter would not fully manipulate the object according to the instruction.

For every training point-cloud/language pair $(p_i, l_i)$, 
we have a set of demonstrations $\mathcal{T}_i$ 
and the most optimal demonstration trajectory $\tau_i^* \in \mathcal{T}_i$.
Using the optimal demonstration $\tau_i^*$ and a loss function $\Delta(\tau, \bar{\tau})$
for comparing demonstrations, 
we find a set of trajectories $\mathcal{T}_{i,S}$ that are relevant (similar)
to this task and a set of trajectories $\mathcal{T}_{i,D}$ that are irrelevant (dissimilar.)
We use the DTW-MT distance function (described later in Sec.~\ref{sec:experiments}) 
as our loss function $\Delta(\tau, \bar{\tau})$,
but it could be replaced by any function that computes the loss of predicting $\bar{\tau}$
when $\tau$ is the correct demonstration.
Using a strategy previously used for handling noise in crowd-sourced data \cite{sung_robobarista_2015},
we can use thresholds $(t_S, t_D)$ to generate two sets from the pool of all trajectories:
$$\mathcal{T}_{i,S} = \{\tau \in \mathcal{T} | \Delta(\tau_i^*, \tau) < t_S \}$$
$$\mathcal{T}_{i,D} = \{\tau \in \mathcal{T} | \Delta(\tau_i^*, \tau) > t_D \}$$

For each pair of $(p_i, l_i)$,
we want all projections of $\tau_j \in \mathcal{T}_{i,S}$ to have higher similarity
to the projection of $(p_i, l_i)$ than $\tau_k \in \mathcal{T}_{i,D}$.
A simple approach would be to train the network
to distinguish these two sets by enforcing
a finite distance (safety margin) between the similarities of
these two sets \cite{weinberger2005distance},
which can be written in the form of a constraint:
$$sim(\Phi_{\mathcal{P},\mathcal{L}}(p_i,l_i), \Phi_{\mathcal{T}}(\tau_j)) 
\geq 1 + sim(\Phi_{\mathcal{P},\mathcal{L}}(p_i,l_i), \Phi_{\mathcal{T}}(\tau_k))$$

Rather than simply being able to distinguish two sets, we want to learn
semantically meaningful embedding spaces from different modalities.
Recalling our earlier example 
where one incorrect trajectory for manipulating
a door knob was much closer to correct than another, it is clear that our
learning algorithm should drive some incorrect trajectories to be more
dissimilar than others.
The difference between the similarities of $\tau_j$ and $\tau_k$ to the projected
point-cloud/language pair $(p_i, l_i)$ should be at least the loss $\Delta(\tau_j,\tau_k)$.
This can be written as a form of a constraint:
\begin{align*}
\forall \tau_j \in \mathcal{T}_{i,S},  \forall \tau_k  \in & \; \mathcal{T}_{i,D} \\
sim(\Phi_{\mathcal{P},\mathcal{L}}(p_i,l_i)&, \Phi_{\mathcal{T}}(\tau_j)) \\
&\geq \Delta(\tau_j, \tau_k) + sim(\Phi_{\mathcal{P},\mathcal{L}}(p_i,l_i), \Phi_{\mathcal{T}}(\tau_k)) 
\end{align*}
Intuitively, this forces trajectories with higher DTW-MT distance from the ground truth to embed further 
than those with lower distance. Enforcing all combinations of these constraints could grow exponentially large.
Instead, similar to the cutting plane method for structural support vector machines \cite{tsochantaridis2005large},
we find the most violating trajectory $\tau' \in \mathcal{T}_{i,D}$
for each training pair of $(p_i, l_i, \tau_i \in \mathcal{T}_{i,S})$
at each iteration.
The most violating trajectory 
has the highest value after the similarity is augmented with the loss scaled by a constant $\alpha$:
$$\tau'_i = \argmax_{\tau \in \mathcal{T}_{i,D}} (sim(\Phi_{\mathcal{P},\mathcal{L}}(p_i,l_i), \Phi_{\mathcal{T}}(\tau)) 
+ \alpha \Delta(\tau_i, \tau))$$

The cost of our deep embedding space $h^3$ is computed 
as the hinge loss of the most violating trajectory.
\begin{align*}
L_{h^3}(p_i, l_i, \tau_i) = |\Delta(\tau_i, \tau'_i) + & sim(\Phi_{\mathcal{P},\mathcal{L}}(p_i,l_i), \Phi_{\mathcal{T}}(\tau'_i))  \\
- & sim(\Phi_{\mathcal{P},\mathcal{L}}(p_i,l_i), \Phi_{\mathcal{T}}(\tau_i))|_+
\end{align*}

The average cost of each minibatch is back-propagated 
through all the layers of the deep neural network
using the AdaDelta \cite{zeiler2012adadelta} algorithm.

\begin{figure}
  \begin{center}
    \includegraphics[width=.95\columnwidth]{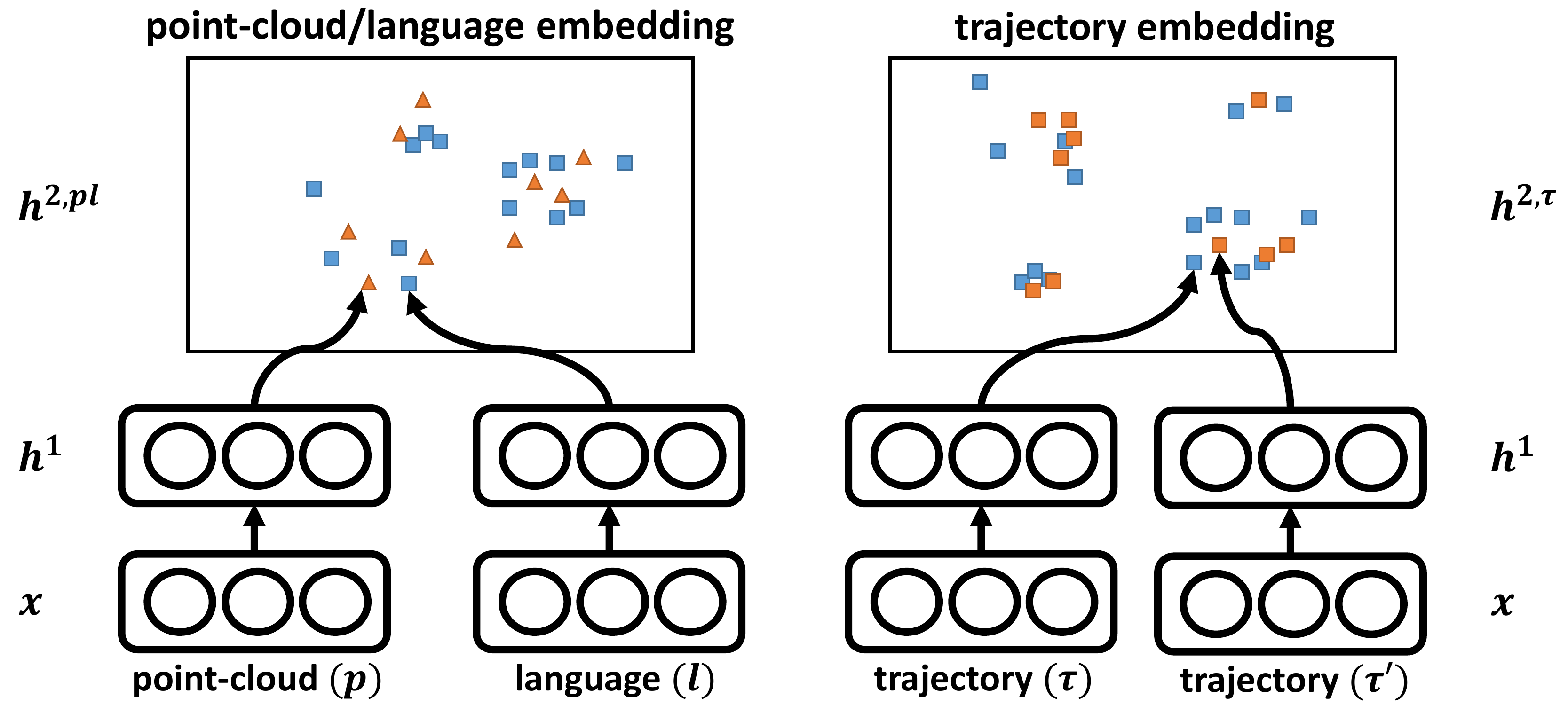}
  \end{center}
  \vskip -.1 in
  \caption{\textbf{Pre-training lower layers:} Visualization of our pre-training approaches for
  $h^{2,pl}$ and $h^{2,\tau}$. For $h^{2,pl}$, our 
  algorithm pushes matching point-clouds and instructions
  to be more similar. For $h^{2,\tau}$, our algorithm
  pushes trajectories with higher DTW-MT similarity to
  be more similar.}
  \label{fig:joint_embedding_pretraining}
  \vskip -.1 in
\end{figure}

\begin{figure*}
  \vskip -0.1in
  \begin{center}
    \includegraphics[width=\textwidth,clip, trim=0cm 0cm 0cm 4cm]{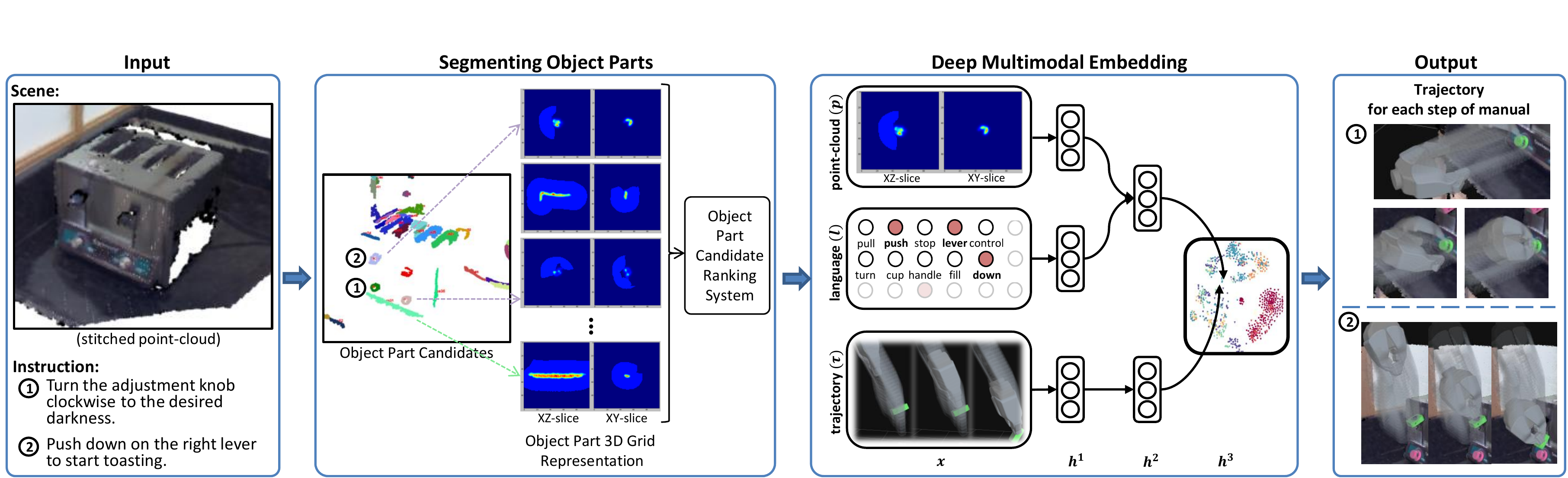}
  \end{center}
  \vskip -0.2in
  \caption{ \textbf{System Overview: }
    Given a point-cloud and a language instruction, our goal is to output a
  	trajectory that would manipulate the object according to the instruction. 
  	The given point-cloud scene is segmented into many parts and ranked for each
  	step of the instruction manual.
  	By embedding point-cloud, language, and trajectory modalities
  	into a joint embedding space, our algorithm selects the best trajectory
  	to transfer to the new object. 
  	} 
  \label{fig:system_overview}
\end{figure*}

\subsection{Pre-training Joint Point-cloud/Language Model}
\label{sec:pre_task}

One major advantage of modern deep learning methods is the use of unsupervised pre-training
to initialize neural network parameters to a good starting point before the final supervised
fine-tuning stage. 
Pre-training helps these high-dimensional networks
to avoid overfitting to the training data.

Our lower layers $h^{2,pl}$ and $h^{2,\tau}$ represent features extracted
exclusively from the combination of point-clouds and language,  
and from trajectories, respectively.
Our pre-training method initializes $h^{2,pl}$ and $h^{2,\tau}$
as semantically meaningful embedding spaces similar to $h^3$,
as shown later in Section~\ref{sec:results}.

First, we pre-train the layers leading up to these layers using 
sparse de-noising autoencoders \cite{vincent2008extracting,zeiler2013rectified}.
Then, our process for pre-training $h^{2,pl}$ is similar to 
our approach to fine-tuning a semantically meaningful embedding space for $h^3$
presented above,
except now we find the most violating language $l'$
while still relying on a loss over the associated optimal trajectory:
$$l' = \argmax_{l \in \mathcal{L}} (sim(\Phi_{\mathcal{P}}(p_i), \Phi_{\mathcal{L}}(l)) + \alpha \Delta(\tau_i, \tau))$$
\begin{align*}
L_{h^{2,pl}}(p_i, l_i, \tau_i) = |\Delta(\tau_i, \tau') + & sim(\Phi_{\mathcal{P}}(p_i), \Phi_{\mathcal{L}}(l'))  \\
- & sim(\Phi_{\mathcal{P}}(p_i), \Phi_{\mathcal{L}}(l_i)) |_+
\end{align*}
Notice that although we are training this embedding space to project from point-cloud/language data, 
we guide learning using trajectory information.

After the projections $\Phi_{\mathcal{P}}$ and $\Phi_{\mathcal{L}}$ are tuned,
the output of these two projections are added to form the output of layer $h^{2,pl}$ in the final feed-forward network.

\subsection{Pre-training Trajectory Model}
\label{sec:pre_traj}

For our task of inferring manipulation trajectories for novel objects, it is especially important
that similar trajectories $\tau$ map to similar regions in the feature space defined by $h^{2,\tau}$, 
so that trajectory embedding $h^{2,\tau}$ itself is semantically meaningful
and they can in turn be mapped to similar regions in $h^3$. 
Standard pretraining methods, such as sparse de-noising autoencoder \cite{vincent2008extracting,zeiler2013rectified}
would only pre-train $h^{2,\tau}$ to reconstruct individual trajectories. 
Instead, we employ pre-training similar to pre-training of $h^{2,pl}$ above, 
except now we pre-train for only a single modality --- trajectory data.

As shown on right hand side of Fig.~\ref{fig:joint_embedding_pretraining},
the layer that embeds to $h^{2,\tau}$ is duplicated.
These layers are treated as if they were two different
modalities,
but all their weights are shared and updated simultaneously. 
For every trajectory $\tau \in \mathcal{T}_{i,S}$, 
we can again find the most violating $\tau' \in \mathcal{T}_{i,D}$ 
and the minimize a similar cost function as we do for $h^{2,pl}$.

\section{Application: Manipulating Novel Objects}
\vspace*{\sectionReduceBot}
\label{sec:systems}

As an example application, we consider manipulating novel appliances \cite{sung_robobarista_2015}.
Our goal is to use our learned embedding
space to allow the robot to infer a manipulation trajectory 
when it is introduced to a new appliance with its natural language instruction manual. 
For example, as shown in Fig.~\ref{fig:system_overview}, 
given a point-cloud of a scene with a toaster and
an instruction such as `Push down on the right lever to start toasting,' 
it should output a trajectory,
representative of how a two-fingered end-effector
should move, including how to approach, grasp, and push down on the lever.
Our algorithm allows the robot to leverage
prior experience with \emph{different} appliances --- for example, a trajectory
which manipulates the handles of a paper
towel dispenser might be transferred to
manipulate the toaster handle.

First, in order to correctly identify a part $p$ out of a scene $s$
that an instruction asks to manipulate,
a point-cloud of a scene $s$ is segmented into many small
potential candidates. 
All segments are ranked for each step of the manual instruction.
Multiple variations of correct segmentations 
and lots of incorrect segmentation 
make our embedding representation even more robust as shown later in Sec.~\ref{sec:results}.

Then, from a library of trajectories with prior experience, 
the trajectory that gives the highest similarity 
to the selected point-cloud $p$ and language $l$ in our embedding space $\mathbb{R}^M$:
$$\argmax_{\tau \in \mathcal{T}} \; sim(\Phi_{\mathcal{P},\mathcal{L}}(p,l), \Phi_{\mathcal{T}}(\tau))$$
As in \cite{weston2011wsabie}, similarity is defined as $sim(a,b) = a \cdot b$.

The previous approach to this problem \cite{sung_robobarista_2015} 
requires projecting a new point-cloud and 
natural language instruction with \emph{every} trajectory in the training set 
exhaustively through the network during inference.

Instead, our approach allows us to pre-embed all candidate trajectories 
into a shared embedding space.
The correct trajectory can be identified 
by embedding only a new point-cloud/language pair.
As shown in Sec.~\ref{sec:results}, this significantly improves 
both the inference run-time and accuracy and 
makes it much more scalable to a larger training dataset.

\subsection{Segmenting Object Parts from Point-clouds}
\label{sec:segmentation}
\vspace*{\subsectionReduceBot}

Our learning algorithm (Sec.~\ref{sec:DME})
assumes that object parts $p$ corresponding to each
instruction $l$ have already been segmented from the point-cloud scene $s$.
While our focus is on learning to manipulate these segmented parts,
we also introduce a segmentation approach which allows us to 
both build an end-to-end system 
and augment our training data for better unsupervised learning,
as shown in Sec.~\ref{sec:results}.

\subsubsection{Generating Object Part Candidates}
We employ a series of geometric feature based techniques
to segment a scene $s$ into small overlapping segments $\{p_1, p_2, ..., p_n\}$.
We first extract Euclidean clusters of points 
while limiting the difference of normals 
between a local and larger region 
\cite{ioannou2012difference,rusu2010semantic}.
We then filter out segments which are 
too big for human hands.
To handle a wide variety of object parts of different scales,
we generate two sets of candidates
with two different sets of parameters, which are combined for evaluation.

\subsubsection{Part Candidate Ranking Algorithm}

Given a set of segmented parts $p$, we must now
use our training data $\mathcal{D}$ to
select for each instruction $l_j$ the
best-matching part $p^*_j$. We do so by optimizing
the score $\psi(p_j,l_j)$ of each segment $p_i$ 
for a step $l_j$ in a manual, evaluated in three parts:
$$\psi(p_i,l_j;\mathcal{D}) = \psi_{feat}(p_i,l_j) \big( \psi_{pc}(p_i,l_j) + \psi_{lang}(p_i,l_j) \big)$$

The score $\psi_{pc}$ is based on the $k_{p}$-most identical segments from the training data $\mathcal{D}$,
based on cosine similarity using our grid representation (Sec.~\ref{sec:representation}).
The score is a sum of similarity against these segments and their associated language:
$\psi_{pc}(p_i,l_j) = \sum_{k=1}^{n} (\text{sim}(p_i,p_k) + \beta\; \text{sim}(l_j,l_k))$.
If the associated language does not exist 
(\emph{i.e.} $p_k$ is not a manipulable part), 
it is given a set similarity value.
Similarly, the score $\psi_{lang}$ is based on the $k_{l}$-most identical language instructions
in $\mathcal{D}$. 
It is a sum of similarity against 
identical language and associated expert point-cloud segmentations. 

The feature score $\psi_{feat}$ is computed by 
weighted features $w^T  \phi_{feat}(p_i,l_j)$ as described in Sec.~\ref{sec:features}.
Each score of the segmented parts $\psi(p_i,l_j)$ is then adjusted 
by multiplying by ratio of its score against the marginalized score in the manual:
$\hat{\psi}(p_i,l_j) = \frac{\psi(p_i,l_j)}{\sum_{l_k \in m_{new}}\psi(p_i,l_k)} \psi(p_i,l_j)$.
For each $l_j \in m_{new}$, 
an optimal segment of the scene chosen as the segment with the maximum score: 
$max_{p_i \in s_{new}} \hat{\psi}(p_i,l_j)$.

\subsubsection{Features}
\label{sec:features}

Three features are computed for each segment in the context of the original scene.
we first infer where a person would stand 
by detecting the `front' of the object,
by a plane segmentation constrained to have a normal axis less than $45^{\circ}$  
from the line between the object's centroid and the original sensor location,
assuming that the robot is introduced close to the `front'.
We then compute a `reach' distance from an imaginary person $170cm$ tall,
which is defined as the 
distance from the head of the person to each segment
subtracted by the distance of the closest one.
Also, because stitched point-clouds have significant noise near their edges,
we compute the distance from the imaginary view ray, a line
defined by the centroid of the scene to the head of the person.
Lastly, objects like a soda fountain a sauce dispenser have many identical parts,
making it difficult to disambiguate different choices (e.g. Coke/Sprite, ketchup/mustard).
Thus, for such scenarios, we also provided a 3D point 
as if human is pointing at the label of the desired selection.
Note that this does not point at the actual part 
but rather at its label or vicinity.
A distance from this point is also used as a feature.


\vspace*{\subsectionReduceTop}
\subsection{Data Representation}
\label{sec:representation}
\vspace*{\subsectionReduceBot}

All three data modalities $(p,l,\tau)$ 
are variable-length and must be transformed into 
a fixed-length representation.


Each point-cloud segment is converted into a real-valued 3D occupancy grid 
where each cell's value is proportional to
how many points fall into the cube it spans.
We use a $100 \times 100 \times 100$ grid of cubic cells with sides of $0.25 cm$.
Unlike our previous work \cite{sung_robobarista_2015},
each cell count is also distributed to the neighboring cells with an exponential distribution.
This smooths out missing points
and increases the amount of information represented.
The grid then is normalized to be between $0 \sim 1$
by dividing by the maximal count.

While our approach focuses on the shape of the part in
question, the shape of the nearby scene can also have
a significant effect on how the part is manipulated.
To account for this, we assign a value of $0.2$ to any
cell which only contains points which belong to the scene
but not the specific part in question, but are within
some distance from the nearest point for the given part.
To fill hollow parts
behind the background, such as tables and walls,
we ray-trace between the starting location of the sensor and 
cells filled by background points and fill these similarly.

While our segment ranking algorithm uses the full-sized grid for each segment,
our main embedding algorithm uses two compact grids generated 
by taking average of cells:
$10 \times 10 \times 10$ grids with cells with sides of $2.5cm$ and of $1cm$.

Natural language $l$ is represented by a fixed-size
bag-of-words.
Trajectories are variable-length sequences of waypoints,
where each waypoint contains 
a position, an orientation and a gripper state
(`open'/`closed'/`holding') defined in the coordinate frame of object part.
All trajectories $\tau$ are then normalized to a fixed length of 
$15$ waypoints. For more details on trajectory representation, 
please refer to \cite{sung_robobarista_2015}.

\vspace*{\subsectionReduceTop}
\subsection{Robotic Platform}
\vspace*{\subsectionReduceBot}

We tested our algorithms on a PR2,
a mobile robot with two arms with seven degrees of freedom each (Fig.~\ref{fig:robotic_exp}).
All software for controlling the robot is written in ROS \cite{quigley2009ros},
and the embedding algorithm are written with Theano \cite{Bastien-Theano-2012}.
All of the computations are done on a remote 
computer utilizing a GPU for our embedding model.

\vspace*{\subsectionReduceTop}
\subsection{Parameters}
\vspace*{\subsectionReduceBot}

Through validation, we found an optimal embedding space size 
$M$ of $25$ and intermediate-layer sizes 
$N_{1,p}$, $N_{1,l}$, $N_{1,\tau}$, $N_{2,pl}$, and $N_{2,\tau}$ of
$250$, $150$, $100$, $125$, and $100$ 
with the loss scaled by $\alpha=0.2$. 
These relatively small layer sizes also had the advantage
of fast inference, as shown in Sec.~\ref{sec:results}.

\section{Experiments}
\label{sec:experiments}
\vspace*{\sectionReduceBot}

In this section, we present two sets of experiments.
First, we show offline experiments on the Robobarista dataset \cite{sung_robobarista_2015} which test 
individual components of our system, showing improvements
for each.
Second, we present a series
of real-world robotic experiments which show that our system is able to produce 
trajectories that can successfully manipulate objects
based on natural language instructions.

\noindent
\textbf{Dataset.}
We test our model on the Robobarista dataset \cite{sung_robobarista_2015}.
This dataset consists of 115 point-cloud scenes with 154 manuals,
consisting of 249 expert segmented point-clouds and 250 free-form natural language instructions.
It also contains 1225 crowd-sourced manipulation trajectories 
which are demonstrated for 250 point-cloud/language pairs.
The point-clouds are collected by stitching multiple views using Kinect Fusion.
The manipulation trajectories are collected from 71 non-experts on Amazon Mechanical Turk.

\noindent
\textbf{Evaluation.}
All algorithms are evaluated using \emph{five-fold cross-validation}, 
with 10\% of the data kept out as a validation set.
For each point-cloud/language pair in test set of each fold,
each algorithm chooses one trajectory from the training set which best suits this pair.
Since our focus is on testing ability to reason about different modalities
and transfer trajectory, segmented parts are
provided as input.
To evaluate transferred trajectories, the dataset contains a separate expert demonstration
for each point-cloud/language pair,
which is not used in the training phase \cite{sung_robobarista_2015}.
Every transferred trajectory is evaluated against these expert demonstrations.

\noindent
\textbf{Metrics.}
For evaluation of trajectories,
we use dynamic time warping 
for manipulation trajectories (DTW-MT)  \cite{sung_robobarista_2015},
which non-linearly warps two trajectories of different lengths 
while preserving weak ordering of matched trajectory waypoints.
Since its values are not intuitive, 
\citet{sung_robobarista_2015} also reports the percentage of transferred trajectories
that have a DTW-MT value of less than 10 from the ground-truth trajectory,
which indicates that it will most likely 
correctly manipulate according to the given instruction
according to an expert survey.



We report three metrics: DTW-MT per manual,
DTW-MT per instruction, and Accuracy (DTW-MT $< 10$) per instruction.
Instruction here refers to every point-cloud/language pair,
and manual refers to list of instructions which comprises a set of sequential tasks, 
which we average over.


\begin{table}[tb]
\vskip -.05in
\begin{center}
\caption{
\textbf{Result on Robobarista dataset} with \emph{5-fold cross-validation}. 
Rows list models we tested including our model and baselines.
Each column shows a different metric used to evaluate the models.
For the DTW-MT metric, lower values are better. For accuracy, higher is better.
}
\begin{tabular}{@{}r|c|c|c@{}}
\hline
 &  \textbf{per manual} & \multicolumn{2}{c}{\textbf{per instruction}} \\ \hline
\textbf{Models}  &  \textbf{DTW-MT}  & \textbf{DTW-MT}  & \textbf{Accu. $(\%)$} \\ \hline
\emph{chance}   & $28.0 \;(\pm 0.8)$ & $27.8 \;(\pm 0.6)$ & $11.2 \;(\pm 1.0)$ \\ \hline
\emph{object part classifier} \cite{sung_robobarista_2015}      & - & $22.9 \;(\pm 2.2)$ & $23.3 \;(\pm 5.1)$ \\ 
\emph{LSSVM + kinematic} \cite{sung_robobarista_2015}    & $17.4 \;(\pm 0.9)$ & $17.5 \;(\pm 1.6)$ & $40.8 \;(\pm 2.5)$\\ 
\emph{similarity + weights} \cite{sung_robobarista_2015}  & $13.3 \;(\pm 1.2)$ & $12.5 \;(\pm 1.2)$ & $53.7 \;(\pm 5.8)$     \\ \hline
\textbf{\emph{Sung et al.} \cite{sung_robobarista_2015}}   & $13.0 \;(\pm 1.3)$ &  $12.2 \;(\pm 1.1)$ &  $60.0 \;(\pm 5.1)$  \\ \hline 

\emph{LMNN-like cost} \cite{weinberger2005distance}      & $15.4 \;(\pm 1.8)$ &  $14.7 \;(\pm 1.6)$ &  $55.5 \;(\pm 5.3)$ \\ \hline    

\textit{Ours w/o pretraining}   & $13.2 \;(\pm 1.4)$ & $12.4 \;(\pm 1.0)$ & $54.2 \;(\pm 6.0)$  \\ \
\textit{Ours with SDA}                 & $11.5 \;(\pm 0.6)$ &  $11.1 \;(\pm 0.6)$ &  $62.6 \;(\pm 5.8)$ \\ \hline

\emph{Ours w/o Mult. Seg}        & $11.0\;(\pm 0.8)$ & $10.5\;(\pm 0.7)$ & $65.1$\; $(\pm 4.9)$  \\ \hline  

\textbf{\emph{Our Model}}        & $\textbf{10.3}\;(\pm 0.8)$ & $\textbf{9.9}\;(\pm 0.5)$ & $\textbf{68.4}$\; $(\pm 5.2)$  \\ \hline  
\end{tabular}
\label{tab:results}
\end{center}
\vskip -.15in
\end{table}


\noindent
\textbf{Baselines.}
We compare our model against several baselines:

1) \textit{Chance:}
Trajectories are randomly transferred.

2) \textit{Sung et al. \citet{sung_robobarista_2015}:} 
State-of-the-art result on this dataset 
that trains a neural network to predict 
how likely each known trajectory matches a given point-cloud and language.

We also report several baselines from this work which rely on more traditional approaches
such as classifying point-clouds into labels like `handle' and `lever' 
(\textit{object part classifier}),
or hand-designing features for multi-modal data 
(\textit{LSSVM + kinematic structure}, \textit{task similarity + weighting}).


3) \textit{LMNN \cite{weinberger2005distance}-like cost function:}
For both fine-tuning and pre-training, 
we define the cost function without loss augmentation.
Similar to LMNN \cite{weinberger2005distance}, 
we give a finite margin between similarities.
For example, as cost function for $h^3$:

\vskip -.15in
{
\small
\begin{align*}
|1 + sim(\Phi_{\mathcal{P},\mathcal{L}}(p_i,l_i), \Phi_{\mathcal{T}}(\tau'))  
- sim(\Phi_{\mathcal{P},\mathcal{L}}(p_i,l_i), \Phi_{\mathcal{T}}(\tau_i))|_+
\end{align*}
}
\vskip -.15in


4) \textit{Our Model without Pretraining:}
Our full model finetuned without any pre-training of lower layers. 

5) \textit{Our Model with SDA:} 
    Instead of pre-training $h^{2,pl}$ and $h^{2,\tau}$
	as defined in Secs.~\ref{sec:pre_task} and \ref{sec:pre_traj},
    we pre-train each layer as stacked de-noising autoencoders 
    \cite{vincent2008extracting,zeiler2013rectified}.

6) \textit{Our Model without Multiple Segmentations:} 
	Our model trained only with expert segmentations,
	without taking utilizing all candidate segmentations in auto-encoders
	and multiple correct segmentations of the same part during training.



\subsection{Results}
\label{sec:results}
\vspace*{\subsectionReduceBot}

\begin{figure}
  \vskip -.05in
  \begin{center}
    \includegraphics[width=.85\columnwidth,trim={2.3cm 9.8cm 2cm 9.8cm},clip]{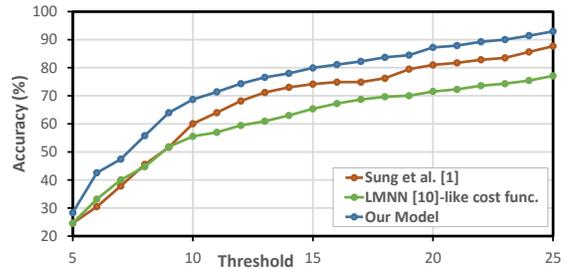}
  \end{center}
  \vskip -.2in
  \caption{
     \textbf{Accuracy-threshold graph} showing results of varying thresholds on DTW-MT measures. Our algorithm consistently
     outperforms the previous approach \cite{sung_robobarista_2015} and an
LMNN-like cost function \cite{weinberger2005distance}. 
     }
  \label{fig:accuracy_plot}
  \vskip -.1in
\end{figure}

We present the results of our algorithm and the baseline approaches in Table~\ref{tab:results}.
Additionally, Fig.~\ref{fig:accuracy_plot} shows accuracies obtained 
by varying the threshold on the DTW-MT measure.

The state-of-the-art result \cite{sung_robobarista_2015} on this dataset has a DTW-MT measure of $13.0$ per manual
and a DTW-MT measure and accuracy of $12.2$ and $60.0\%$ per instruction.
Our full model based on joint embedding of multimodal data
achieved $10.3$, $9.9$, and $68.4\%$, respectively.
This means that when a robot encounters a \emph{new} object it has never seen
before, our model gives a trajectory which would correctly manipulate it
according to a given instruction approximately $68.4\%$ of the time. 
From Fig.~\ref{fig:accuracy_plot}, we can see
that our algorithm consistently outperforms both prior work and an LMNN-like cost
function for all thresholds on the DTW-MT metric.

\begin{figure*}[tb]
  \vspace*{-0.05in}
  \begin{center}
    \includegraphics[width=\textwidth,height=3.5in,trim={0cm 0cm 0cm 3.2cm},clip]{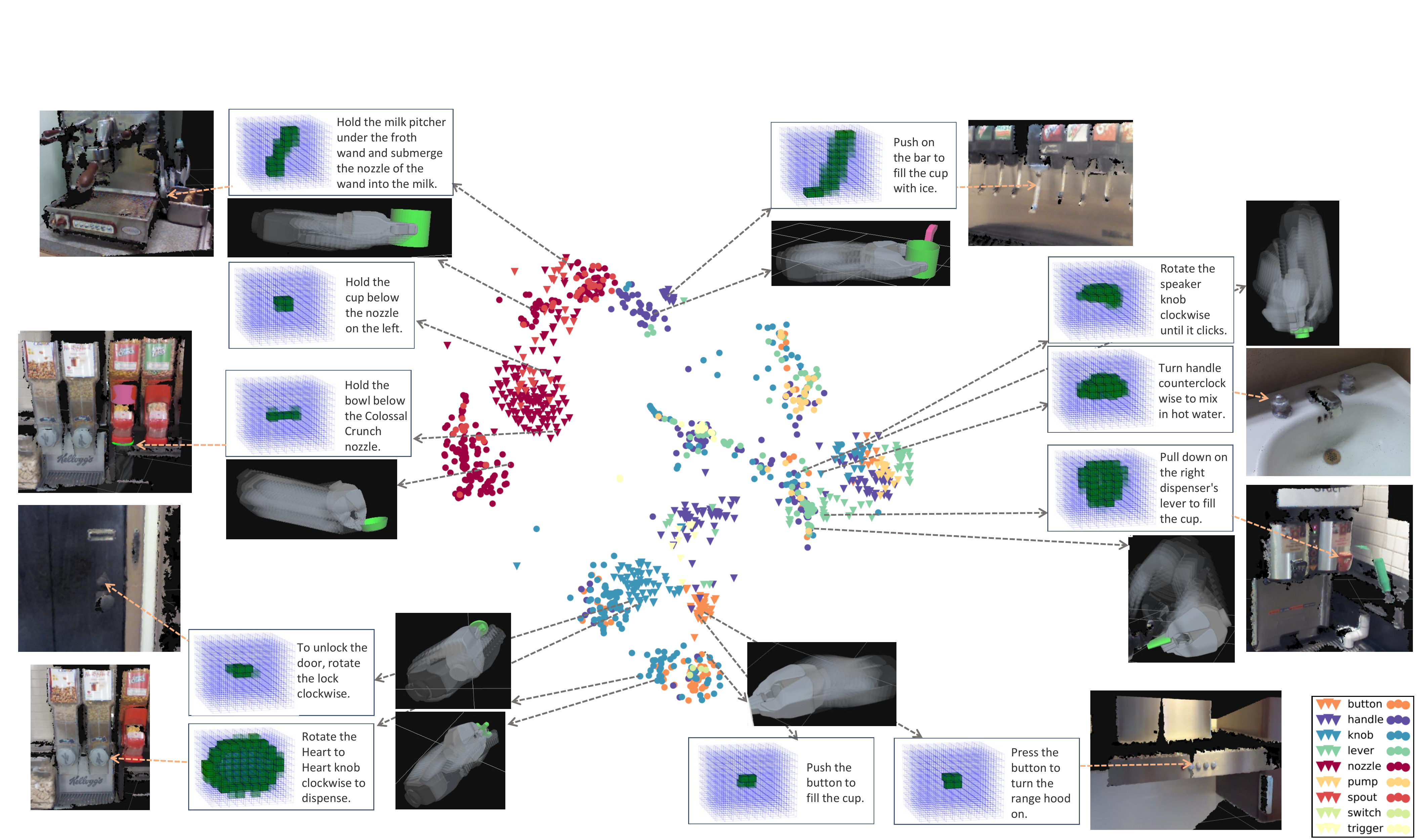}
  \end{center}
  \vskip -.23in
  \caption{
    \textbf{Learned Deep Point-cloud/Language/Trajectory Embedding Space: }
     Joint embedding space $h^3$ after the network is fully fine-tuned, 
     visualized in 2d using t-SNE \cite{van2008visualizing} .
     Inverted triangles represent projected point-cloud/language pairs, circles represent projected trajectories. 
     The occupancy grid representation (Sec.~\ref{sec:representation})
     of object part point-clouds is shown 
     in \textcolor{darkgreen}{green} in \textcolor{blue}{blue} grids.
     For presentation purpose, `neighbor' cells are not shown.
     The legend at the bottom right shows classifications of object parts by an expert, 
     collected for the purpose of building a baseline.
     As shown by result of this baseline (object part classifier in Table~\ref{tab:results}), 
     these labels do not necessarily correlate well with the actual manipulation motion.
     Thus, full separation according to the labels defined in the legend is not optimal and
     will not occur.
     These figures are best viewed in color. 
     }
  \label{fig:embed_vis}
    \vskip -.18in

\end{figure*}

\textbf{What does our learned deep embedding space represent?}
Fig.~\ref{fig:embed_vis} shows a visualization of the top layer $h^3$, the joint embedding space.
This visualization is created by projecting all training data (point-cloud/language pairs and trajectories) 
of one of the cross-validation folds to $h^3$, 
then embedding them to 2-dimensional space 
using t-SNE \cite{van2008visualizing}.

One interesting result is that our system was able to naturally learn that ``nozzle'' and ``spout''
are effectively synonyms for purposes of manipulation. It clustered these 
together in the upper-left of Fig.~\ref{fig:embed_vis} based solely on the
fact that both are associated with similar point-cloud shapes and manipulation
trajectories. 

In addition to the aforementioned cluster,
we see several other logical clusters. 
Importantly, we can see that our embedding maps vertical
and horizontal rotation operations to very different 
regions of the space --- roughly 2 o'clock and 8 o'clock
in Fig.~\ref{fig:embed_vis}, respectively.
Despite the fact that
these have nearly identical language instructions, our
algorithm learns to map them differently based on their
point-clouds, mapping nearby the appropriate manipulation
trajectories.

\begin{figure}[t]
  \vskip -.05in
  \begin{center}
    \includegraphics[width=.9\columnwidth,trim={0cm 1cm 7cm 1cm},clip]{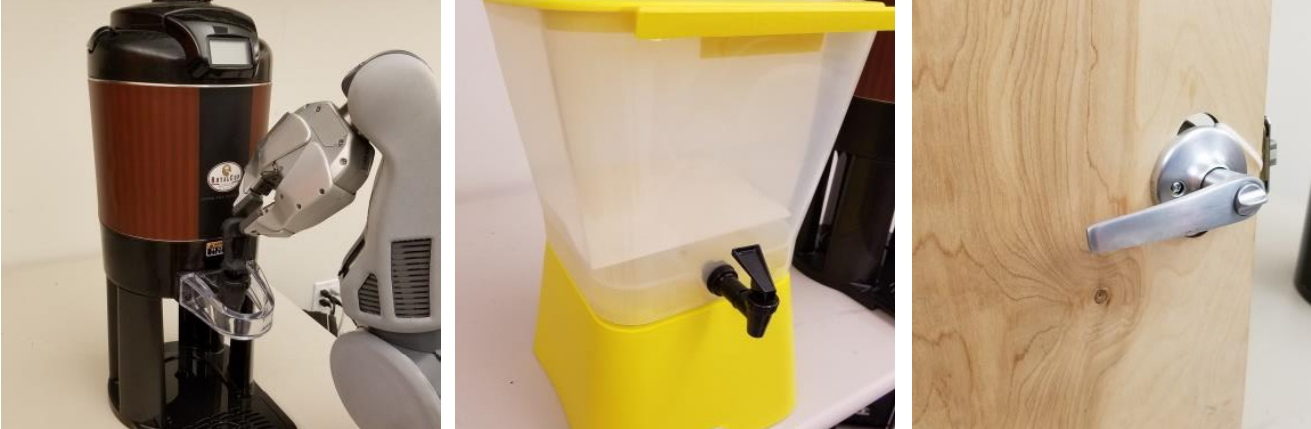}
  \end{center}
  \vskip -.2in
  \caption{
    \textbf{Robotic Experiments:} 
    We test our algorithm on a PR2 robot with 
    three different novel objects --- coffee dispenser handle, beverage dispenser lever, and door handle.
     }
  \label{fig:robotic_exp}
  \vskip -.15in
\end{figure}

\textbf{Should the cost function be loss-augmented?}
When we change the cost function for pre-training $h^2$ and fine-tuning $h^3$
to use a constant margin of $1$ between relevant $\mathcal{T}_{i,S}$ and 
irrelvant $\mathcal{T}_{i,D}$ demonstrations \cite{weinberger2005distance},
performance drops to $55.5\%$.
This loss-augmentation is also visible in our embedding space.
Notice the purple cluster around the 12 o'clock region of Fig.~\ref{fig:embed_vis},
and the right portion of the red cluster in the 11 o'clock region.
The purple cluster represents tasks and demonstrations related to pushing a bar,
and the lower part of the red cluster represents the task of holding a cup below the nozzle.
Although the motion required for one task would not be replaceable by the other,
the motions and shapes are very similar, especially compared to 
most other motions
e.g. turning a horizontal knob.

\textbf{Is pre-embedding important?}
As seen in Table~\ref{tab:results}, without any pre-training
our model gives an accuracy of only $54.2\%$. Pre-training
the lower layers with the conventional stacked 
de-noising auto-encoder (SDA) algorithm \cite{vincent2008extracting,zeiler2013rectified}
increases performance to $62.6\%$, still significantly
underperforming our pre-training algorithm, 
at $68.4\%$. 
This shows that our metric embedding pre-training approach provides a better
initialization for an embedding space than SDA.

\textbf{Can automatically segmented object parts be manipulated?}
From Table~\ref{tab:robotic_result}, we see that
our segmentation approach was able to find 
a good segmentation for the object parts in question
in $50$ of $60$ robotic trials (Sec.~\ref{sec:robotic_experiment}), or 83.3\% of the time. 
Most failures occurred for the beverage dispenser, which had a small lever that was difficult to segment. 

When our full DME model utilizes two variations of same part
and uses all candidates as a training data for the auto-encoder,
our model performs at $68.4\%$ compared to $65.1\%$
which only used expert segmentations.

\textbf{Does embedding improve efficiency?}
While \citet{sung_robobarista_2015} has $749,638$ parameters to be learned, 
our model only has $616,175$ (and still gives better performance.)

The previous model had to compute joint point-cloud/language/trajectory features 
for all combinations of the current point-cloud/language pair with \emph{each} candidate trajectory
(i.e. all trajectories in the training set) to infer an optimal trajectory. 
This is inefficient and does not scale well with the number of training
datapoints.
However, our model pre-computes the projection of all trajectories into $h^3$.
Inference in our model then requires only projecting
the new point-cloud/language combination to $h^3$ once and 
finding the trajectory with maximal similarity in this embedding.

In practice, this results in a significant improvement in efficiency, 
decreasing the average time to infer a trajectory
from $2.3206$ms to $0.0135$ms, a speed-up of about $171$x.
Time was measured on the same hardware, with a GPU (GeForce GTX Titan X), 
using Theano \cite{Bastien-Theano-2012}.
We measured inference times $10000$ times for first test fold, which has a pool of $962$ trajectories.
Times to preprocess the data and load into GPU memory were not included in this measurement.

\begin{table}[tb]
\vskip -.05in
\begin{center}
\caption{
\textbf{Results} of $60$ experiments on a PR2 robot
running end-to-end experiments autonomously on three different objects.}

\vskip -.02in
\begin{tabular}{@{}r|c|c|c|c@{}}
\hline
\textbf{Success Rate} &  \textbf{Dispenser}  & \textbf{Beverage}  & \textbf{Door} &   \\
\textbf{of Each Step}  &  \textbf{Handle}  & \textbf{Lever}  & \textbf{Handle}  &\textbf{Avg.}  \\
\hline
\emph{1) Segmentation}             & $90.0\%$  & $65.0\%$  & $95.0\%$  & $83.3\%$  \\ \hline 
\emph{2) DME Traj. Inference} & $94.4\%$  & $100.0\%$ & $78.9\%$  & $91.1\%$  \\ \hline 
\emph{3) Execution of Traj.}  & $82.4\%$  & $76.9\%$  & $100.0\%$ & $86.4\%$  \\ \hline 
\end{tabular}

\label{tab:robotic_result}
\end{center}
\vskip -.3in
\end{table}

\vspace*{-0.04in}
\subsection{Robotic Experiments}
\label{sec:robotic_experiment}
\vspace*{\subsectionReduceBot}

To test our framework, we performed $60$
experiments on a PR2 robot 
in three different scenes shown in Fig.~\ref{fig:robotic_exp}.
We presented the robot with the object placed within reach
from different starting locations along with a language instruction.

Table~\ref{tab:robotic_result}
shows results of our robotic experiments.
It was able to successfully complete the task end-to-end autonomously $39$ times.
Segmentation was not as reliable for the beverage dispenser
which has a small lever.
However, when segmentation was successful,
our embedding algorithm was able to provide a correct
trajectory with an accuracy of $91.1\%$.
Our PR2 was then able to correctly follow these trajectories
with a few occasional slips due to the relatively
large size of its hand compared to the objects.

Video of robotic experiments are available at this website: 
\url{http://www.robobarista.org/dme/}



\vspace*{\sectionReduceTop}
\section{Conclusion}
\vspace*{\sectionReduceBot}

In this work, we introduce an algorithm that learns a semantically meaningful
common embedding space
for three modalities --- point-cloud, natural language and trajectory.
Using a loss-augmented cost function, we learn to embed 
in a joint space such that similarity of any two points in the space 
reflects how relevant to each other.
As one of application, we test our algorithm on the problem of manipulating novel
objects.
We empirically show on a large dataset that our embedding-based approach 
significantly improve accuracy, 
despite having less number of learned parameters and being
much more computationally efficient (about $171$x faster) than the 
state-of-the-art result.
We also show via series of robotic experiments that our segmentation algorithm
and embedding algorithm allows robot to autonomously perform the task.
\noindent
\textbf{Acknowledgment.}
This work was supported 
by Microsoft Faculty Fellowship and NSF Career Award to Saxena.

\bibliographystyle{IEEEtran}
\bibliography{writeup}



\end{document}